\documentclass[graybox]{svmult}

\usepackage{mathptmx}       
\usepackage{helvet}         
\usepackage{courier}        
\usepackage{type1cm}        
\usepackage{makeidx}         
\usepackage{graphicx}        
\usepackage{multicol}        
\usepackage[bottom]{footmisc}

\makeindex             

\begin{document}

\title*{Data-Driven Dialogue Systems for Social Agents}
\titlerunning{Data-Driven Dialogue Systems for Social Agents}
\author{{\bf Kevin K. Bowden, Shereen Oraby, Amita Misra, Jiaqi Wu, and Stephanie Lukin} \\
Natural Language and Dialogue Systems Lab \\ 
University of California, Santa Cruz \\
\{kkbowden, soraby, amisra2, jwu64, slukin\}@ucsc.edu}
\authorrunning{K. Bowden, S. Oraby, A. Misra, J. Wu, and S. Lukin} 

\date{}

%
%

\maketitle

\abstract{In
  order to build dialogue systems to tackle the ambitious task of
  holding social conversations, we argue that we need a data-driven approach that includes insight into human conversational ``chit-chat'', and which
  incorporates different natural language processing modules. Our
  strategy is to analyze and index large corpora of social media data,
  including Twitter conversations, online debates, dialogues
  between friends, and blog posts, and then to couple this data
  retrieval with modules that perform tasks such as sentiment and
  style analysis, topic modeling, and summarization. We aim for
  personal assistants that can learn more nuanced human language, and
  to grow from task-oriented agents to more personable social
  bots.  }
  
\section{From Task-Oriented Agents to Social Bots}
Devices like the Amazon Echo and Google Home have entered our homes to perform task-oriented functions,
such as looking up today's headlines and setting reminders
\cite{echo,google}. As these devices evolve, we have begun to expect social conversation, where the device must learn to personalize and produce natural language style.

Social conversation is not explicitly goal-driven in the same way as task-oriented dialogue. Many dialogue systems in both the written and spoken medium have been developed for task-oriented agents with an explicit goal of restaurant information retrieval, booking a flight, diagnosing an IT issue, or providing automotive customer support \cite{hirschman2000evaluating,price1992subject,walker1997paradise,walker2001quantitative,Henderson2014,Tsiakoulis12}. 
These tasks often revolve around question answering, with little ``chit-chat''.
Templates are often used for generation and state tracking, but since
they are optimized for the task at hand, the conversation can 
either become stale, or maintaining a conversation requires the 
intractable task of  manually authoring many different social interactions
 that can be used in a particular context.

We argue that a social agent should be
spontaneous, and allow for human-friendly conversations that do not follow a perfectly-defined trajectory. In
order to build such a conversational dialogue system, we exploit the abundance of human-human social media
conversations, and develop methods informed by natural language
processing modules that model, analyze, and generate utterances that better suit the context. 

\section{Data-Driven Models of Human Language}

A myriad of social media data has led to the development of new techniques
for language understanding from open domain conversations, and many
corpora are available for building data-driven dialogue systems
\cite{Ritter2011,Serban2015}.
%
While there are differences between how people speak in person and in an online text-based environment, the social agents we build should not be limited in their language; they should be exposed to many different styles and vocabularies. Online conversations can be repurposed in new dialogues, but only if they can be properly indexed or adapted to the context. Data retrieval algorithms have been successfully employed to
co-construct an unfolding narrative between the user and computer
\cite{swanson2008say}, and re-use existing conversations \cite{Higashinaka14}. Other approaches train on such
conversations to analyze sequence and word patterns, but lack detailed
annotations and analysis, such as emotion and humor
\cite{sordoni2015neural,vinyals2015neural,li2016persona}. The large Ubuntu Dialogue Corpus \cite{DBLP:conf/sigdial/LowePSP15} with over 7 million utterances is large enough to train neural network models \cite{DBLP:journals/corr/KadlecSK15, DBLP:journals/dad/LowePSCLP17}.

We argue that combining data-driven retrieval with modules for
sentiment analysis and style, topic analysis, summarization,
paraphrasing, and rephrasing will allow for more human-like social
conversation. This requires that data be indexed based on domain and
requirement, and then retrieve candidate utterances based on dialogue
state and context. Likewise, in order to avoid stale and repetitive utterances, we
can alter and repurpose the candidate utterances; for
example, we can use paraphrase or summarization to create new ways of
saying the same thing, or to select utterance candidates
according to the desired sentiment
\cite{Misraetal_naacl15,Misraetal_sigdial16}. The style of
an utterance can be altered based on requirements; introducing
elements of sarcasm, or aspects of factual and emotional argumentation
styles \cite{oraby_am15,oraby_sigdial16}. Changes in the perceived
speaker personality can also make more
personable conversations \cite{lukin_games14}. Even utterances from
monologic texts can be leveraged by converting the content to dialogic
flow, and performing stylistic
transformations \cite{bowden_icids16}.

Of course, while many data sources may be of interest for indexing knowledge
for a dialogue system, annotations are not always available or easy to obtain.
By using machine learning models designed to classify different classes of interest,
such as sentiment, sarcasm, and topic, data can be bootstrapped to greatly increase
the amount of data available for indexing and utterance selection \cite{oraby_am15}.

There is no shortage of human generated dialogues, but the challenge is
to analyze and harness them appropriately for social-dialogue
generation. We aim to combine data-driven methods to repurpose
existing social media dialogues, in addition to a suite of tools for
sentiment analysis, topic identification, summarization, paraphrase,
and rephrasing, to  develop a socially-adept
agent that can carry on a natural conversation.

\bibliographystyle{plain}
\bibliography{author}

\end{document}